\documentclass{article}

\usepackage{arxiv}

\usepackage[utf8]{inputenc} 
\usepackage[T1]{fontenc}    
\usepackage{hyperref}       
\usepackage{url}            
\usepackage{booktabs}       
\usepackage{amsfonts}       
\usepackage{nicefrac}       
\usepackage{microtype}      
\usepackage{lipsum}		
\usepackage{graphicx}
\usepackage{colortbl}
\usepackage{multirow}
\usepackage{tabularx}

\usepackage{color}

\title{General DeepLCP model for disease prediction : Case of Lung Cancer}

\date{}	

\author{ {\hspace{1mm}Mayssa Ben~Kahla}\thanks{} \\
	University of Sousse, Higher Institute of Applied Science and Technology, 4003,
Sousse, Tunisia\\
	\texttt{mayssabenkahla7@gmail.com} \\
	\And
	\hspace{1mm}Dalel ~kanzari \\
	University of Sousse, Higher Institute of Applied Science and Technology, 4003,
Sousse, Tunisia\\
	University of Manouba, National School of Computer Sciences, RIADI Laboratory
2010, Manouba, Tunisia\\
	\texttt{kndalel@yahoo.fr} \\
\And
	\hspace{1mm}Ahmed ~Maalel \\
	University of Sousse, Higher Institute of Applied Science and Technology, 4003,
Sousse, Tunisia\\
	University of Manouba, National School of Computer Sciences, RIADI Laboratory
2010, Manouba, Tunisia\\
	\texttt{maalel.ahmed@gmail.com} \\
}

\renewcommand{\shorttitle}{General DeepLCP model for disease prediction : Case of Lung Cancer}

\hypersetup{
pdftitle={General DeepLCP model for disease prediction : Case of Lung Cancer},
pdfsubject={},
pdfauthor={M.ben Kahla,D.kanzari,A.Maalel},
pdfkeywords={Deeplearning,  NLP, Disease, Prevention,Lung Cancer,CNN},
}

\begin{document}
\maketitle

\begin{abstract}
	 According to GHO (Global Health Observatory (GHO), the high prevalence of a large variety of  diseases such as Ischaemic heart disease, stroke, lung cancer disease and lower respiratory infections have remained the top killers during the past decade.

The growth in the number of mortalities caused by these disease is due to the very delayed symptoms'detection.
Since in the early stages, the symptoms are insignificant and similar to those of benign diseases (e.g. the flu ), and we can only detect the disease at an advanced stage. 

In addition, The high frequency of improper practices that are harmful to health, the hereditary factors, and the stressful living conditions can increase the death rates.

Many researches dealt with these fatal disease, and most of them applied advantage machine learning models to deal with image diagnosis. However the  drawback is that imagery permit only to detect disease at a very delayed stage
and then patient can hardly be saved.

In this Paper we present our new approach "DeepLCP" to predict fatal diseases that threaten people's lives. It's  mainly based on raw and heterogeneous data of the concerned (or under-tested) person. "DeepLCP" results of a combination combination of the Natural Language Processing (NLP) and the  deep learning paradigm.The experimental results of the proposed model in the case of Lung cancer prediction have approved high accuracy and a low loss data rate during the validation of the disease prediction.
\end{abstract}


\section{ Introduction}
       According to GHO Ischaemic heart disease, stroke, lung cancer disease and lower respiratory infections have remained the top killers during the past decade. They cause increasing numbers of deaths worldwide.
For example, Lung cancer is the fourth most common cancer in france according to the national cancer institue with 49109 new cases estimated in 2017 (32260 men and 16849 women)\footnote{"institut national du cancer," accélérons les progrés aux cancers : faces aux cancers .,https://www.e-cancer.fr/Professionnels-de-sante/Les-chiffres-du-cancer-en-France/Epidemiologie-des-cancers/Les-cancers-les-plus-frequents/Cancer-du-poumon., 17-01-2018. Accessed: 28- 06-2019.}, also lung cancer is the leading cause of cancer death in Canada,according to the Canadian cancer society.28600 Canadians were diagnosed with lung cancer, wich represents 14\% of all new cancer cases in 2017.21100 Canadians died from lung cancer, accounting for 26\% of all deaths in 2017\footnote{"institut national du cancer,", accélérons les progrés aux cancers : faces aux cancers .,https://www.e-cancer.fr/Professionnels-de-sante/Les-chiffres-du-cancer-en-France/Epidemiologie-des-cancers/Les-cancers-les-plus-frequents/Cancer-du-poumon., 17-01-2018. Accessed: 28- 06-2019.}.

The growth in the number of mortalities caused by these fatal diseases is generally due to the very delayed detection of symptoms.

The objective of our approach is to estimate the probability of having disease by the deal of person' general life condition, the clinical \textbf{symptoms} such as « chest pain », « persistent cough », etc, and\textbf{ risk factors} like « the family history of the disease » and « smoking », etc.. To achieve our goal we we propose a new approach "DeepLCP" that  combines the natural language processing and the deep learning paradigm to highlight hidden information with significant impact on the disease. 

\section{Related Works}

There are many works that used Deep Learning in the field of medicine. In the approach\cite{Sergey}, the authors used deep learning in structural and functional brain imaging data. They used the Deep Belief Network (DBN) and the Boltzmann machine building block (RMB) to describe  large data dimensions. The problem of this approach is quadratic time complexity.Talathi \cite{Talathi} proposed an in-depth learning framework through the use of Managed Recurring Units (GRUs: Gated Recurrent textbfUnit) for crisis detection. The proposed method  can detect about 98\% of epileptic seizures in the first 5 seconds of the overall duration of the seizure. According to Webb's \cite{sarah} study, Steve Finkbeiner's lab  used a convolutional neuron network (CNN)  to identify , the dead neurons in a living cell population. The weakness of this work  is that it can no longer control the classification process and precisely explain the software outputs .Rajpurkar et al \cite{Rajpurkar}  used a convolutional network architecture (CNN) for musculoskeletal abnormalities detection, in particular the elbow  abnormalities, forearm, hand humerus, shoulder, and wrist from musculoskeletal X-ray data set.
Also,  a lot of works are interested in deep learning for cancer detection. For example the study of  Gruetzemacher and Gupta \cite{Gruetzemacher},  used  revolutionary image recognition method and DeepLearning methods, to distinguish between large and small pulmonary nodules potentially malignant lung nodules. Esteva and his colleagues \cite{Esteva} proposed a classification of skin lesions using a single CNN from biopsy images to detect cancer. 
In the article\cite{Park}, the authors used a DeepLearning algorithm to predict the presence or absence of lung cancer in a chest x-ray by using twelve thousand cases of biopsically proved lung cancer.
Yang and al \cite{Yang1} research, implemented  a DCNN-based classification system to detect lung cancer.
The study of Bychkov and its authors \cite{Dmitrii} consisted of combining convolutional(CNN) and recurent(RNN) architectures to form a deep network and predict colorectal cancer results from images of tumor tissue samples.
In the field of general medicine many workers combined deep learning with natural language processing. For example the work of Hughes and his colleagues\cite{Hughes},  introduced a new approach to classify at the level of sentences, the medical documents.They  showed that it is possible to use CNNs to represent the semantics of the clinical text allowing semantic classification at the level of the sentence. The approach of \cite{Baker}, consisted of a convolutive neural network (CNN) with a biomedical classification of texts to identify the characteristics of cancer such as "Sustainable proliferative signaling",and "Resistance to cell death". The authors of this work based their CNN architecture on the simple model of kim\cite{kim}. Qui and his colleagues \cite{JohnX},  studied deep learning and in particular the convolutional neural network (CNN), to extract ICDO-3 topographic codes from a corpus of cancer pathology reports breast and lung. 

We summarize in Table 1 the different works dealing with deep learning, in general medicine, and in cancer detection.

 In Table 1 we classify also the different works according to their input’s type.
\begin{table}
\centering
\tiny
\arrayrulecolor{black}
\begin{tabularx}{\textwidth}{|X|X|X|X|X|X|X|}
\hline
 Reference       &   \textit{Algorithm}  & \textit{Architecture} & \textit{Input Data}& \textit{Goal} &  \textit{Accuracy} & \textit{Error rate} \\
\hline
~\centering\cite{Sergey} &  Sigmoïde/ SVM /The error return propagation algorithm/Iterative algorithm "divide and compete" (DC)                & DBN / RBM (Unsupervised) &  \begin{center} MRI image\end{center}&\begin{center} Detection of Brain Disease\end{center} & \begin{center} 90\%\end{center}  &\begin{center} \_  \end{center}               \\ 
\hline
~\centering \cite{Talathi}& Back propagation through time (BPTT)/ Logistic regression
                                                                              &  \begin{center}GRU-RNN \end{center}&   \begin{center} EEG recording image\end{center} &\begin{center} Detection of Epileptic Seizures\end{center}&\begin{center}99.6\% \end{center}    &\begin{center} \_ \end{center} \\
\hline

~\centering\cite{sarah}& CNN (Deep thoughts)
                                                                                & \begin{center} CNN (supervised)\end{center}& \begin{center} annotated  image\end{center}&\begin{center}Detection of \textbf{Dead neurons}\end{center}&\begin{center} 90\% -99\%\end{center}
     &\begin{center} \_ \end{center}  \\
\hline
~\centering \cite{Rajpurkar}&\begin{center}\_  \end{center}
                                                                                & \begin{center} CNN \end{center} & \begin{center} X-ray image\end{center}&\begin{center} Detection of Musculoskeletal abnormalities \end{center}&\begin{center} 95\%\end{center} 
     &\begin{center} \_  \end{center}\\
\hline

~\centering\cite{Gruetzemacher} & maxpooling, softmax
                                                                             & \begin{center} DNN \end{center} &  \begin{center}CT scan image\end{center}&\begin{center} Classification of the Characteristics of \textbf{malignant lung nodule}\end{center}&\begin{center} 81.08\% -82.10\%\end{center}
     &\begin{center} \_  \end{center}\\
\hline
~\centering  \cite{Esteva}& Partitioning algorithm (PA)/ Inference algorithm /t-SNE(t-Distributed Stochastic Neighbor Embedding)
                                                                              &  \begin{center}CNN (GoogleNet) (Supervised)\end{center}
 & \begin{center}Biopsy image\end{center}&\begin{center}Classification  of Skin lesions\end{center}&\begin{center}55.4\% -72.1\%\end{center}  &\begin{center}\_ \end{center} \\
\hline

~\centering 
\cite{Park}& Genetic algorithm/ DeepNEAT-Dx/ Retropropagation/ SchiffMan encoding
                                                                              & \begin{center} CNN \end{center}
 & \begin{center} X-ray image\end{center}&\begin{center}Prediction of Lung Cancer\end{center}&\begin{center}96.00\% \end{center}

     &\begin{center} 7.97\% \end{center} \\
\hline
~\centering   \cite{Yang1}& \begin{center}\_ \end{center}
                                                                            & \centering DCNN
 & \begin{center} CT scan image\end{center}&\begin{center}Detection of Lung Cancer\end{center}&\begin{center} 90\%-97\%\end{center} 

     &\begin{center}10\%-24\% \end{center} \\
\hline

~\centering \cite{Dmitrii} & SVM/naive Bayesian classification/ Logistic regression
                                                                            & \begin{center} CNN-RNN  (LSTM) (Supervised)\end{center}
 &\begin{center}Images of tissue microarray (TMA)\end{center}&\begin{center} Prediction of Colorectal Cancer\end{center}&\begin{center} 69\% \end{center}

     &\begin{center} \_ \end{center}\\
\hline
~\centering  \cite{Hughes} & NLP-WORD2VEC
                                                                              & \begin{center} CNN \end{center}
 &\begin{center}Clinical Data\end{center}&\begin{center}Classification of  General Disease\end{center}&\begin{center} 68\% \end{center}   &\begin{center} \_ \end{center} \\
\hline

~\centering \cite{Baker} & NLP-WORD2VEC/ Word embeddings
                                                                              & \begin{center} CNN \end{center}
 &\begin{center}Clinical Data\end{center}&\begin{center}Classification of Characteristics of Cancer\end{center}&\begin{center}  97.1\% \end{center}   &\begin{center} \_ \end{center} \\
\hline

~\centering \cite{JohnX} & Descente adaptative en gradient d'Adadelta/NLP-WORD2VEC /N-grams

                                                                              & \begin{center}CNN \end{center}
 &  \begin{center}Clinical Data\end{center}&\begin{center}Detection of  Breast and Lung Cancer\end{center}&\begin{center}71\% \end{center}   &\begin{center} 30\% \end{center} \\
\hline
\end{tabularx}
\caption{Synthesis Of Related Works}\label{tab1}
\arrayrulecolor{black}
\end{table}
According to Table 1 we note that despite the advanced related works and their significant contributions we remarques certain limits:
\begin{itemize}
\item [$\bullet$ ] {The first weakness is related to the performance degree of these models to detect disease which is lower than those depicted from radiographic image analysis.}
\item [$\bullet$ ] {The second  limitation is related to the weakness of precision in the work that has combined NLP with CNN in the field of medicine.}
\item [$\bullet$ ] {The Third limit is related to the training time, which is too long.}
\item [$\bullet$ ] {Also, most of the related works based their research on the clinical data and don't emphasize the patient's quality of life.}

\end{itemize}
We remark that few approaches use text as an input and most of them use deep learning model after X-ray image analysis. Unfortunately these techniques have the limit to detect the desease at an advanced stage.

To address these issues, we propose the "DeepLCP" model, which is the combination of Natural Language Processing (NLP) and the deep learning concept to prevent fatal disease  by estimating the probability of having the disease.

\section{DeepLCP Workflows}
 Our "DeepLCP" model is mainly composed of two sub-processes, as shown in Figure1: \textbf{ Data Preprocessing} and \textbf{Data Deep processing}.
\begin{itemize}
\item [$\bullet$ ]{ \textbf{Data Preprocessing} is  represented by the data extraction  and the data cleaning .}  
\item [$\bullet$ ]{\textbf{Data Deep processing} is made up of three process: semantic transformation, semantic classification,  and deeplearning algorithm.}
\end{itemize}
Both processing proceed in operational flow in order to predict the probability of having  the disease.
\begin{figure}
\includegraphics[height=5cm,width=17cm]{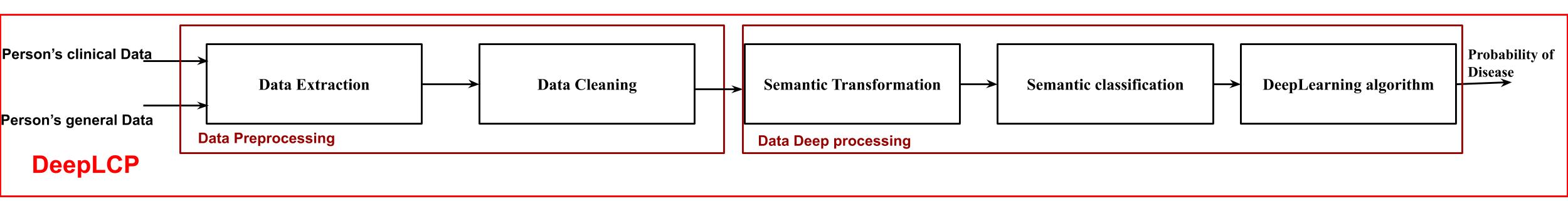}
\caption{General DeepLCP Model}
\end{figure}
\subsection{Data Preprocessing}
 \subsubsection{Data Extraction}
\begin{figure}
\includegraphics[height=7cm,width=17cm]{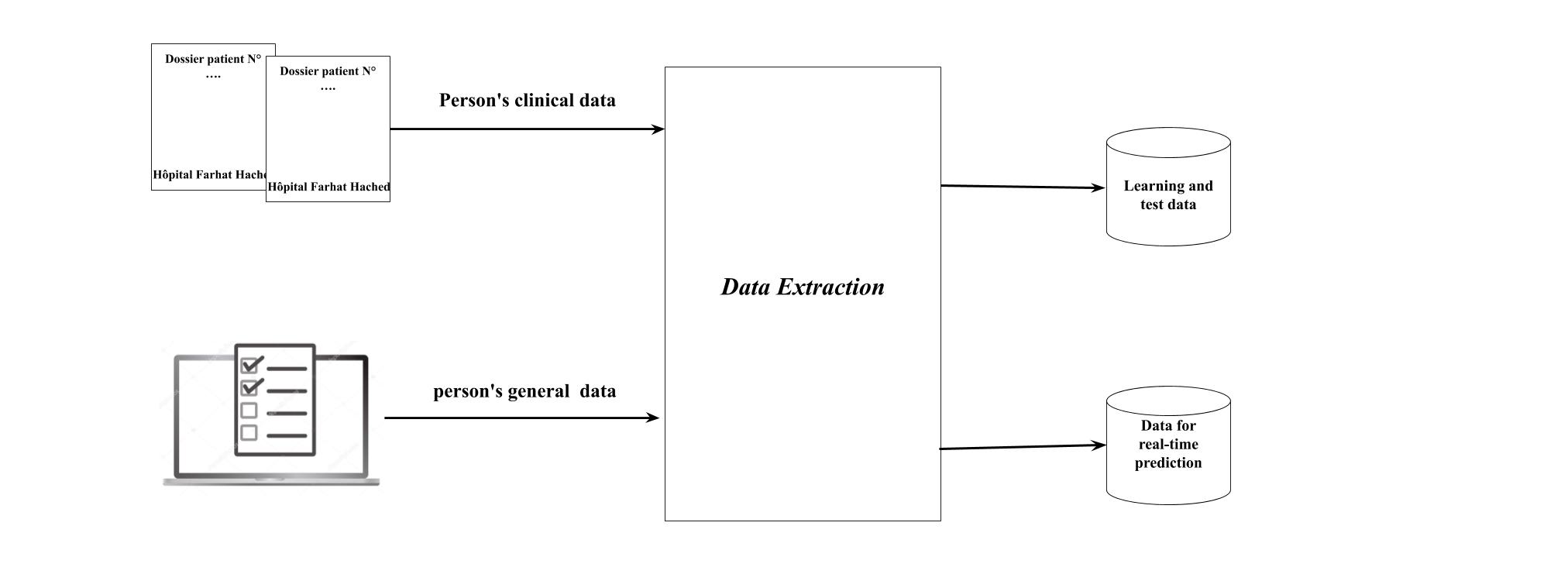}
\caption{Data Extraction}
\end{figure}
As shown in Figure 2, in the data extraction, we collect two type of data :
\begin{itemize}
\item [$\bullet$ ] {\textbf{Person's Clinical Data :}represents data of regarding patient symptoms.}
\item [$\bullet$ ]{\textbf{Person's General Data:}represents data on major and minor risk factors for the disease.}
\end{itemize}

\subsubsection{Data Cleaning}
In this stage, we consider three methods of data cleaning :
\begin{itemize}
\item [$\bullet$ ] {\textbf{Deleting the irrelevant Data: }we  use this method to treat data that doesn't actually fit the specific problem.It's based on the suggestion of experts.}
\item [$\bullet$ ]{\textbf{Fix typos Data:}we use this method when the strings can be entered in many different ways, and  mistakes can be made. }
\item [$\bullet$ ]{\textbf{Standardize Data:}in this method, our data is to not only recognize the typos but also put each value in the same standardized format. E.g. for strings, we have to make sure that all values are either in lower or uppercase.}
\end{itemize}
The data cleaning  help to eliminate the inaccurate information that may lead to bad decision making.

\subsection{Data Deep Processing}
The core of the  DeepLCP architecture is composed of two parts: \textbf{NLP (Natural language processing)}  and Deeplearning algorithm as shown in Figure 3.
\begin{figure}
\includegraphics[height=7cm,width=17cm]{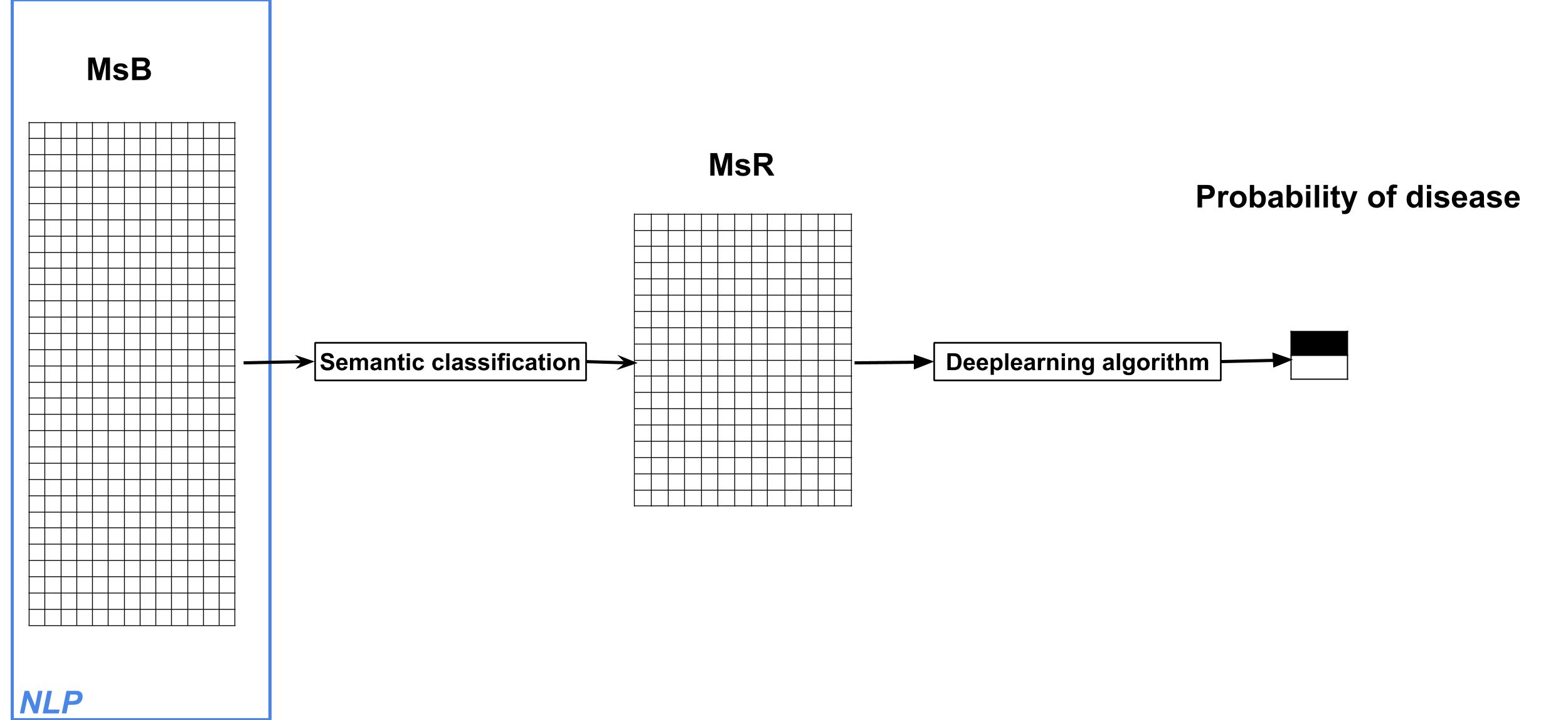}
\caption{ DeepLCP Architecture } \label{fig3}
\end{figure}
\subsubsection{Semantic Transformation}
In the semantic transformation process, we use the natural language processing \textbf{(NLP)} in particular, the word2vec model to convert the sentences into a vector. The latter is generated, according to the user's response and \textbf{the rules of semantic transformations}.

The output of the semantic transformation is the raw semantic matrix \textbf{(MsB)} as illustrated in figure 4.
\begin{figure}
\centering
\includegraphics[height=7cm,width=17cm]{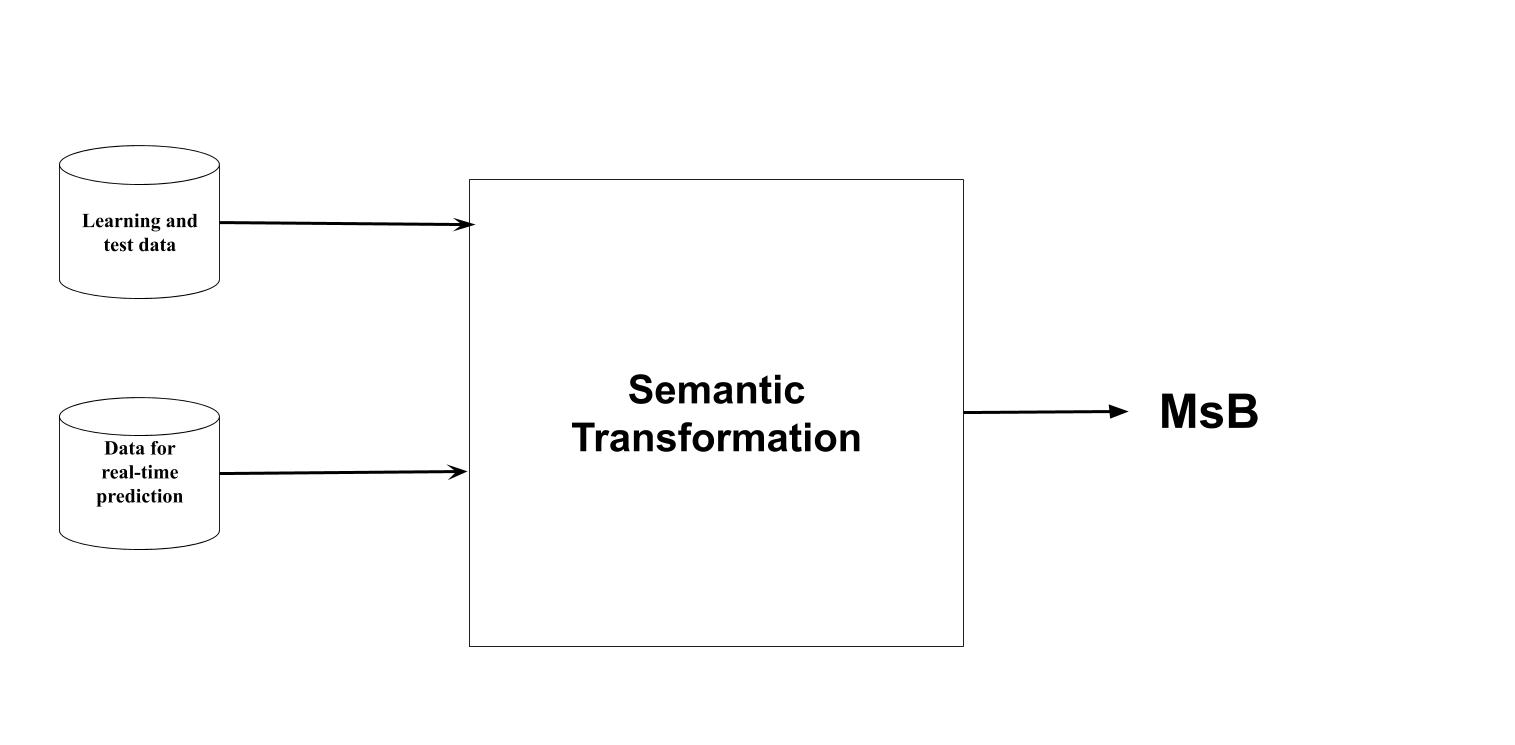}
\caption{Semantic Transformation.} \label{fig4}
\end{figure}

The semantic transformation converts each row in the database into a raw Semantic Matrix  output. Each entry of the \textbf{raw Semantic Matrix (MsB)} represents an individual clinical observation.
\subsubsection{Semantic classification}
From the \textbf{MsB}, the semantic classification makes it possible to generate the \textbf{reduced semantic matrix (MsR)} with a restricted number of rows. The purpose of this section is to reveal the symptoms and major risks of the disease. Figure 5 gives a general overview of the semantic classification.

\begin{figure}
\centering
\includegraphics[height=5cm,width=15cm]{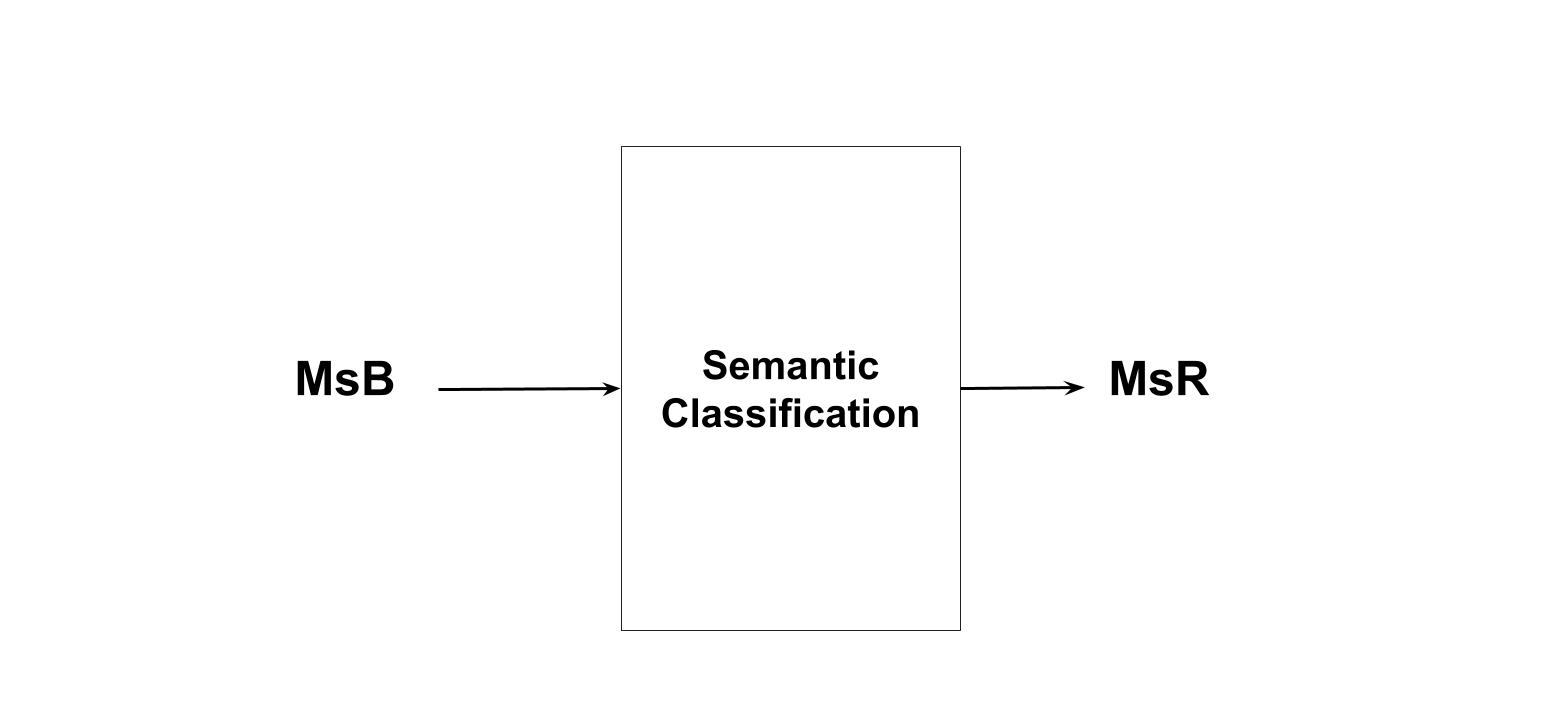}
\caption{Semantic Classification.} \label{fig5}
\end{figure}

\subsubsection{DeepLearning Algorithm}
In this process,we apply deeplearning algorithm on the semantic matrix reduced in order to predict the probability of having disease.
our objective is to extract features which characterize the disease and from this features the model calculate the probability of disease.

\section{Expriment}
We instantiation Deep LCP model in experimentation by using the convolutional neural network (CNN) in the case of lung Cancer prevention. The figure 6 represent DeepLCP model in the case of lung cancer. 

\begin{figure}
\includegraphics[height=3cm,width=15cm]{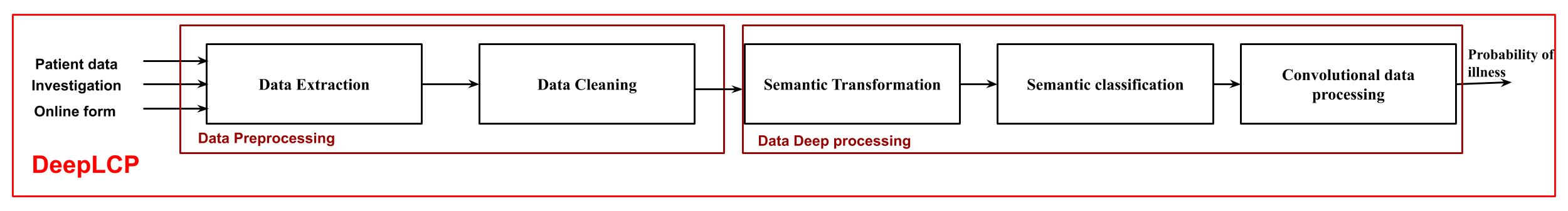}
\caption{DeepLCP Model}
\end{figure}
\subsection{Data Extraction}
As shown in Figure 6, in the data extraction, we collect the data from three different sources:
\begin{itemize}
\item [$\bullet$ ] {the first source is the archived patient files of the Farhat Hached Hospital.}
\item [$\bullet$ ]{the second source is our online survey of non-sick individuals.}
\item [$\bullet$ ]{the third source is our online form of individuals wishing clinical examination of the disease.}
\end{itemize}
The data extraction from the first and the second source contains informal descriptive data of the individuals under examination. An individual is represented by a set of characteristics such as age, gender, profession, etc.

Our database is fed from the archive files of patients with lung cancer, Farhat Hached Hospital in Sousse. We consulted and seized 355 records of patients with the disease. For the unaffected, we launched an online survey  and then we were able to collect 246 descriptions. All data are archived in a csv file.

The data is subdivided into two subsets:
\begin{itemize}
\item [$\bullet$ ] {a learning set: we use 490 real cases, divided into 315 of lung cancer patients and 175 who are unaffected with lung cancer.}
\item [$\bullet$ ] {a test set: we use 111 real cases, divided into 40 patients affected by lung cancer and 71 cases not affected by the disease.}
\end{itemize}
\subsection{Semantic transformation}
as shown in section 3.2.1, the part of the semantic transformation is generating, according to the user's response and the rules of semantic transformations.
\begin{itemize}
\item[$\bullet$ ]{\textbf{Semantic Transformations Rules:}}
We use the formal language Z\cite{refz} to represent 31 semantic transformation rules. We use these rules to associate each piece of information with an incidence weight. E.g. in figure 7 we propose rule 1:
\begin{itemize}
\item [$\bullet$ ] {\textbf{Rule 1 (Figure 7)}: This rule permits to give a weight for the vector associated with gender information. The intervals of weight are suggested by the doctor "Pr.Bouaouina Noureddine", the chef of the radiotherapy department at the hospital Farhat hached Sousse and the doctor "Dr.jalel Knani", Pulmonologist at Tahar Sfar Hospital.according to the two doctors men have more risk of having the disease than women.}
\end{itemize}
\end{itemize}
\begin{figure}
\includegraphics[height=2cm,width=15cm]{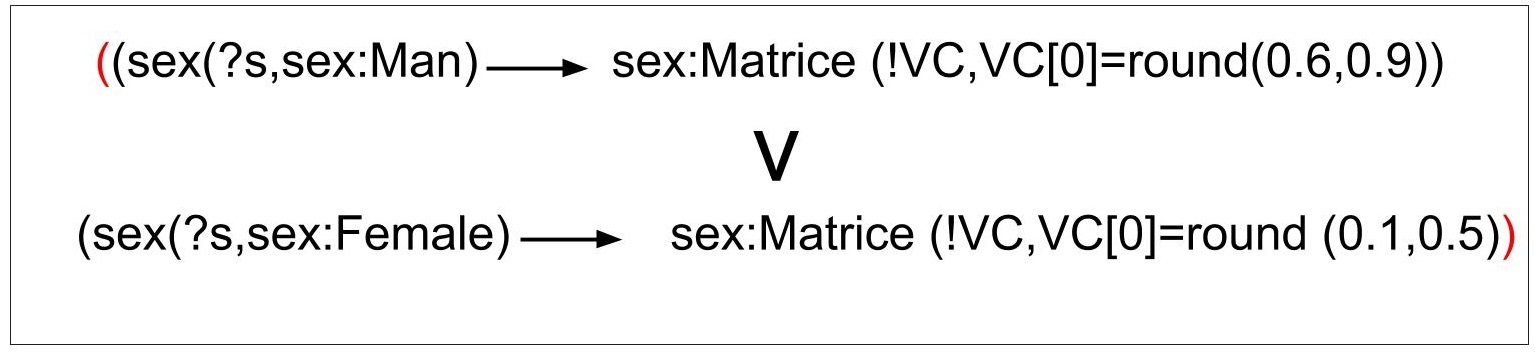}
\caption{Rule 1 of transformation in Z language.} \label{fig6}
\end{figure}

The output of the semantic transformation part is the raw semantic matrix as illustrated in figure 8. The raw semantic matrix has size of [31 * 13].
\begin{itemize}
\item [$\bullet$ ] {31:represents the number of sentences (characteristics) representing an individual.}
\item [$\bullet$ ] {13:represents the maximum number of words in the longest sentence.}
\end{itemize}
\begin{figure}
\centering
\includegraphics[height=8cm,width=23cm]{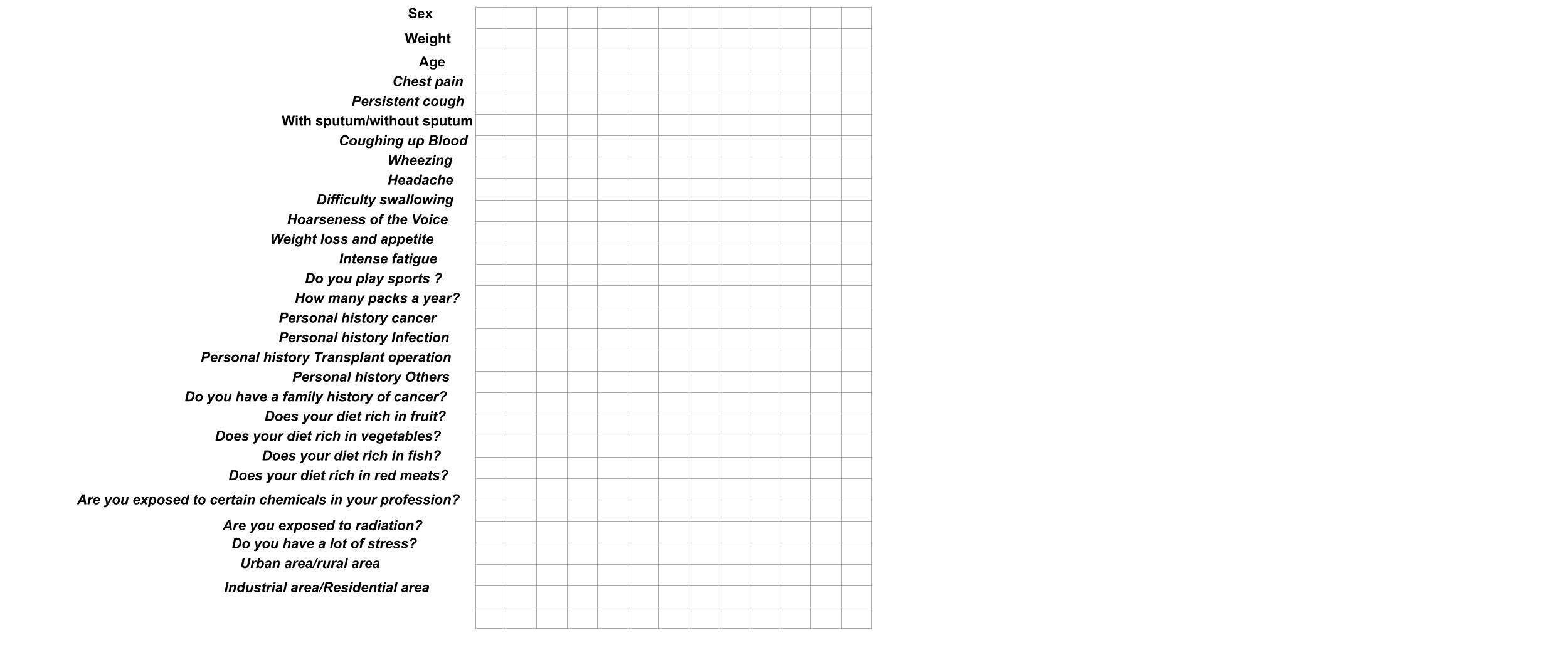}
\caption{Raw Matrix.} \label{fig7}
\end{figure}
\subsection{Semantic Classification}

The purpose of this part is to extract the symptoms, minor and major risks factors of lung cancer disease. In the semantic classification part we apply two types of classification on the raw semantic matrix in order to obtain the reduced semantic matrix:
\begin{itemize}
\item [$\bullet$ ] {\textbf{Classification by categories}: classify the data according to three classes: \textbf{ minor risk factors, major risk factors, and symptoms}. These three categories are the main factors of lung cancer.}
\item [$\bullet$ ] {\textbf{Classification by theme}: classify the data according to six themes: \textbf{Thoracic signs, Cough, Feeding, Consumer, Personal history, Residence.}}

After classification, we obtain the semantic matrix reduced with size [18 * 13]. We use the classification to optimize the number of features. As illustrated in figure 9.

\begin{figure}
\includegraphics[height=7.5cm,width=17cm]{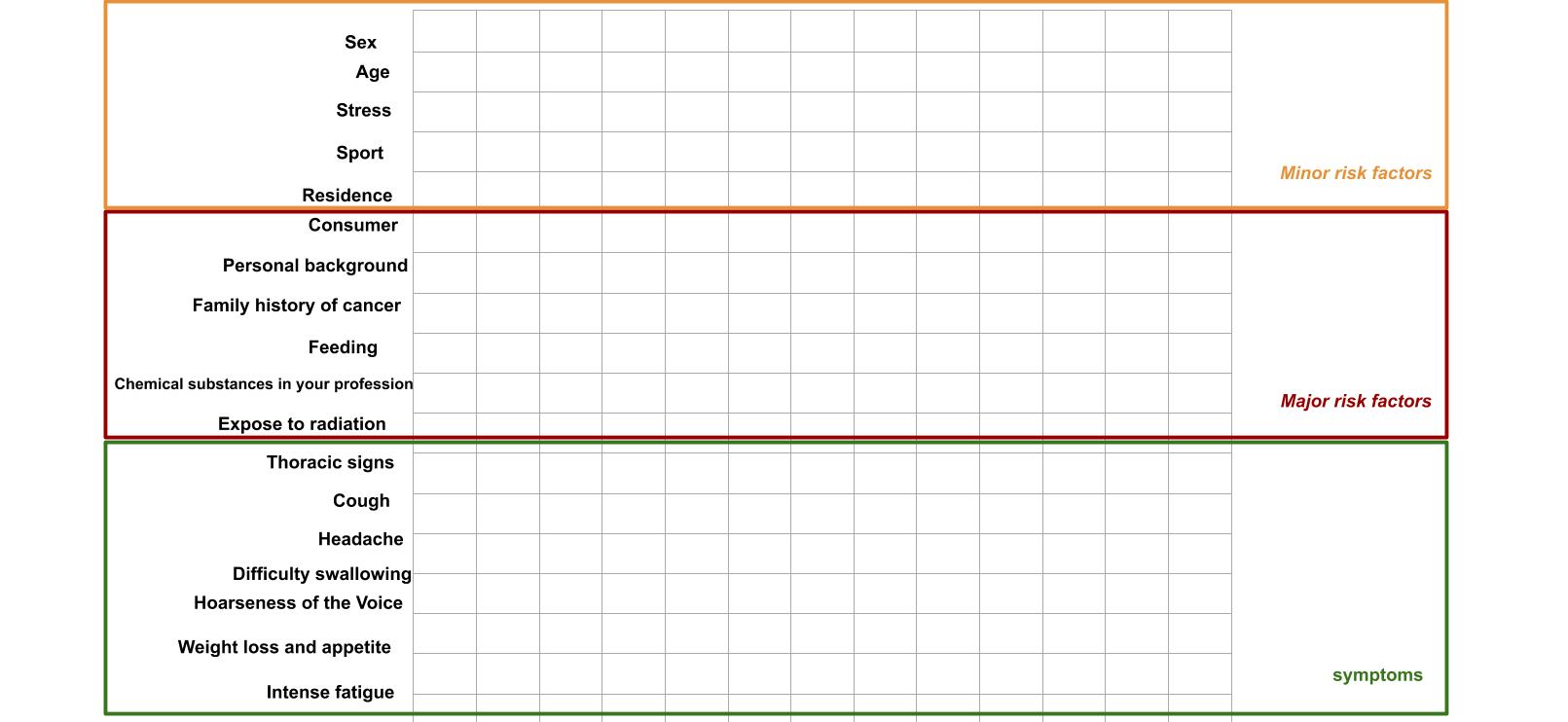}
\caption{ Semantic Matrix Reduced.} \label{fig8}
\end{figure}
\end{itemize}

\begin{itemize}
\item [$\bullet$ ] {The reduction will not lose any information.}
\end{itemize}

\subsection{Convolution Data Processing}

 Our CNN architecture composed of a convolution layer, a pooling layer, and a fully connected layer.

As illustred in Figure 10, in the convolution layer we have 3 filter region sizes: 5, 6, 7. The region size 5 corresponds to the part of "minor risk factors", the region size 6 corresponds to the part of "factors of major risks ", the region size 7 corresponds to the part of" Symptoms ". Each region contains 2 filters. So in total we have 6 filters in our architecture. All filters will scan the semantic matrix reduced, which represents our input with 1 stride to give us a feature map for each filter. We set the "stride" with 1 to extract all the characteristics of the matrix.

Then in the pooling layer, we apply the maxpooling for each feature map to obtain a maximum value for each feature map. Then we concatenate the 6 values obtained in the maxpooling layer.

Finally, in the fully connected layer, we apply the softmax activation function on its concatenated values to obtain our output which composes of two probabilities:
\begin{itemize}
\item [$\bullet$ ] {The probability of having the disease.}
\item [$\bullet$ ] {The probability of not having lung cancer.}
\end{itemize}

\begin{figure}
\includegraphics[height=4cm,width=15cm]{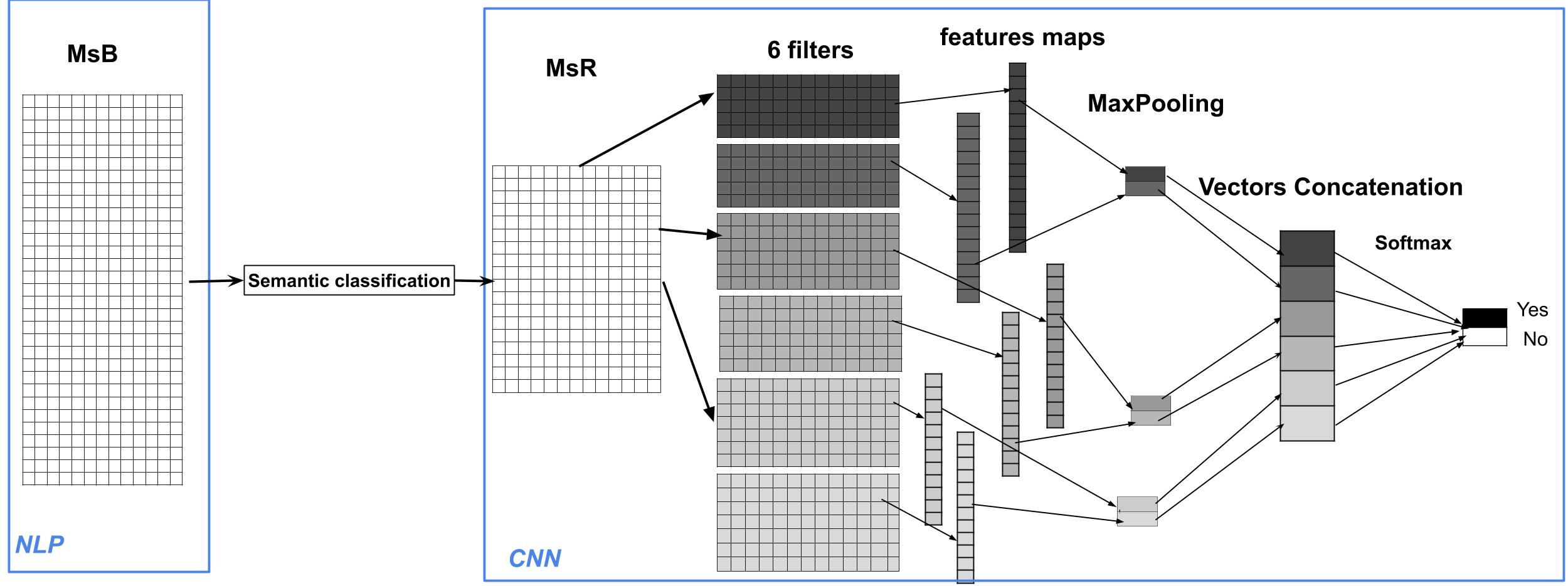}
\caption{Architecture DeepLCP } \label{fig9}
\end{figure}

\subsection{Results obtained from our model}

We obtaine the simulation results shown in Figure 11 We notice that the validation test is 94.59\% and the train precision is 93.88\% for the value of loss in the test validation 0.1699 and for the train is 0.1773.
\begin{figure}
\includegraphics[height=4.5cm,width=10cm]{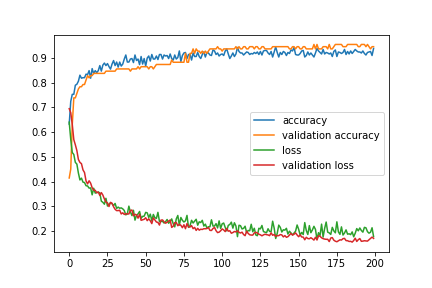}
\caption{Result of the accuracy and loss values for the train part and the test part} \label{fig10}
\end{figure}

From the \textbf{AUC} curve (\textbf{A}rea \textbf{U}nder the \textbf{ROC}(\textbf{R}eceiver \textbf{O}perating \textbf{C}haracteristic) \textbf{C}urve) shown in Figure 12 we find that our model achieves a considerable performance value with a very high true positive rate and our classification is strong. The value of AUC is 0.99 (very close to 1) means that it has a good measure of separability. The AUC is one of the most important evaluation parameters to verify the performance of a classification model.
\begin{figure}
\includegraphics[height=4.5cm,width=10cm]{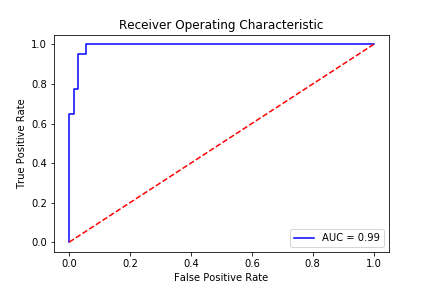}
\caption{Result of the accuracy and loss values for the train part and the test part} \label{fig11}
\end{figure}
\section{Discussion}
We test our dataset with four machine learning algorithm:

\begin{itemize}
\item [$\bullet$ ] {First, we test our database with the\textbf{ k-nearest neighbors (KNN)} algorithm with neighborhood values = 5 with this type of algorithm we obtain 86.48\% as precision value and 13.52\% error rate.}
\item [$\bullet$ ] {Second, we test the proposed dataset with \textbf{Decision Tree} with this algorithm we obtain  93.69\% as precision value and as 6.31\% error rate.}
\item [$\bullet$ ] {Third,  we  test with the \textbf{Random Forest algorithm} with this algorithm we obtain 91.89\% as precision value and 8.11\% as Error rate.}
\item [$\bullet$ ] {Finally,  we test the model with the \textbf{ANN \cite{dalel}} algorithm in this algorithm we use 18 nodes in the Input layer and 10 nodes in the hidden layer and a 'relu' activation function, then the flatten layer, to avoid the overfittings we use dropout = 0.75 with an output layer of 2. For this model we obtain 85.59\% as precision value and 14.41\% as error rate after 200 epochs.}
\end{itemize}
\section*{Conclusion}
In this article, we present a new generic model for the prevention and detection of a fatal disease. Our model called "DeepLCP"  consists mainly of a combination of two approaches the natural language processing(NLP) and the deeplearning paradigm. Our model is based on semantic transformation. In the exprimentation we instantiate "DeepLCP" model by using formal transformation rules and  the convolutional neural network (CNN).The accuracy of the validation test is 94.5\%, which confirms that our model gives an effective result.

As perspectives, we plan to enrich our model by using the incremental learning algorithms.
\bibliographystyle{model1-num-names}

\end{document}